%% file: main.tex
\documentclass[10pt,twocolumn,letterpaper]{article}

\usepackage{cvpr}
\usepackage{times}
\usepackage{epsfig}
\usepackage{graphicx}
\usepackage{amsmath}
\usepackage{amssymb}
\usepackage{booktabs}

\usepackage[dvipsnames]{xcolor}
\usepackage{algorithm}
\usepackage[noend]{algpseudocode}
\usepackage{multirow}

\usepackage{tabularx}
\usepackage{xcolor}

\usepackage[pagebackref=true,breaklinks=true,letterpaper=true,colorlinks,bookmarks=false]{hyperref}
\usepackage{pifont}% http://ctan.org/pkg/pifont

\cvprfinalcopy % *** Uncomment this line for the final submission

\pdfoutput=1

\def\cvprPaperID{2963} % *** Enter the CVPR Paper ID here

\definecolor{silver}{rgb}{0.95,0.95,0.95}

\ifcvprfinal\fi

\newcolumntype{C}{>{\centering\arraybackslash}X}
\begin{document}

%%%%%%%%%%%%%%%%%%
% our commands
\newcommand{\rk}[1]{\textcolor{CornflowerBlue}{[D: #1]}}
\newcommand{\move}[1]{\textcolor{DarkOrchid}{[Should be moved] #1}}
\newcommand{\myparagraph}[1]{\vspace{0.1em}\noindent {\bf #1.}}
\newcommand{\todo}[1]{\textcolor{red}{[TODO:#1]}}
\newcommand{\alert}[1]{\textcolor{red}{[#1]}}
\newcommand{\jsays}[1]{\textcolor{JungleGreen}{[Jon says: #1]}}
\newcommand{\cmark}{\ding{51}}%
\newcommand{\xmark}{\ding{55}}%
\newcommand{\red}[1]{\textcolor{red}{\textbf{#1}}}
\newcommand{\gray}[1]{\textcolor{gray}{\textbf{#1}}}
\newcommand{\cyan}[1]{\textcolor{cyan}{#1}}
\newcommand{\pink}[1]{\textcolor{pink}{#1}}

\makeatletter
\def\BState{\State\hskip-\ALG@thistlm}
\makeatother
\def\NoNumber#1{{\def\alglinenumber##1{}\State #1}\addtocounter{ALG@line}{-1}}

%%%%%%%%% TITLE
\title{Re-ID done right: towards good practices for person re-identification}

\author{
Jon Almaz\'an$^{1}$ ~ ~ Bojana Gaji\'c$^2$\thanks{Work done during an internship at NAVER LABS Europe.} ~ ~ Naila Murray$^1$ ~ ~ Diane Larlus$^1$ \\
\centering
\begin{minipage}{.4\textwidth}
\centering
$^1$\small{Computer Vision Group\\NAVER LABS Europe\\}
{\tt\small firstname.lastname@naverlabs.com} 
\end{minipage}
\begin{minipage}{.4\textwidth}
\centering
$^2$\small{Computer Vision Center\\Dept. de Ci\`encies de la Computaci\'o, UAB\\}
{\tt\small bgajic@cvc.uab.es} 
\end{minipage} 
}

\maketitle
%\thispagestyle{empty}

%%%%%%%%% ABSTRACT
\begin{abstract}

Training a deep architecture using a ranking loss has become standard for the person re-identification task.
Increasingly, these deep architectures include additional components that leverage part detections, attribute predictions, pose estimators and other auxiliary information, in order to more effectively localize and align discriminative image regions.
In this paper we adopt a different approach and carefully design each component of a simple deep architecture and, critically, the strategy for training it effectively for person re-identification.
We extensively evaluate each design choice, leading to a list of good practices for person re-identification.
By following these practices, our approach outperforms the state of the art, including more complex methods with auxiliary components, by large margins on four benchmark datasets.
We also provide a qualitative analysis of our trained representation which indicates that, while compact, it is able to capture information from localized and discriminative regions, in a manner akin to an implicit attention mechanism. 
\end{abstract}

%%%%%%%%% BODY TEXT
\input{introduction}

\input{relatedwork}

\input{method}

\input{experiments}

\input{conclusions}

%\newpage
{\small
\bibliographystyle{ieee}
\bibliography{biblio}
}
\end{document}

%% file: introduction.tex
\section{Introduction}
\label{sec:intro}

Person re-identification (re-ID) is the task of correctly identifying individuals across different images captured under varying conditions, such as different cameras within a network.
This task is of high practical value in a wide range of applications including surveillance or content-based image retrieval.
Different from classification, there is no overlap between the persons seen at train time and at test time.

Heavily studied for more than two decades \cite{bedagkar2014survey, karanam2016comprehensive}, most works that address this problem have sought to propose either a suitable image representation, often with hand-crafted rules, or a suitable image similarity metric.
Following the great success of deep learning in a large number of computer vision tasks, including image classification \cite{he16deep}, object detection \cite{ren2015faster}, and semantic segmentation \cite{dai2016instance}, a dominant paradigm in person re-ID has emerged, where methods use or fine-tune successful deep architectures for this retrieval task \cite{chen17beyond, hermans17indefense, su16deep}.
This paradigm leads to compact global image representations well-suited for person re-identification.
However, within this general framework there remain many design choices, in particular those related to network architectures, training data, and model training, that have a large impact on the effectiveness of the final person re-ID model.
%Yet, this type of approaches varies by a large number of design choices that seem to have sometimes a large impact on the final results \cite{}. 
In this paper, we focus on identifying which of these design choices matter.

\begin{figure}[t!]
  \includegraphics[width=\linewidth]{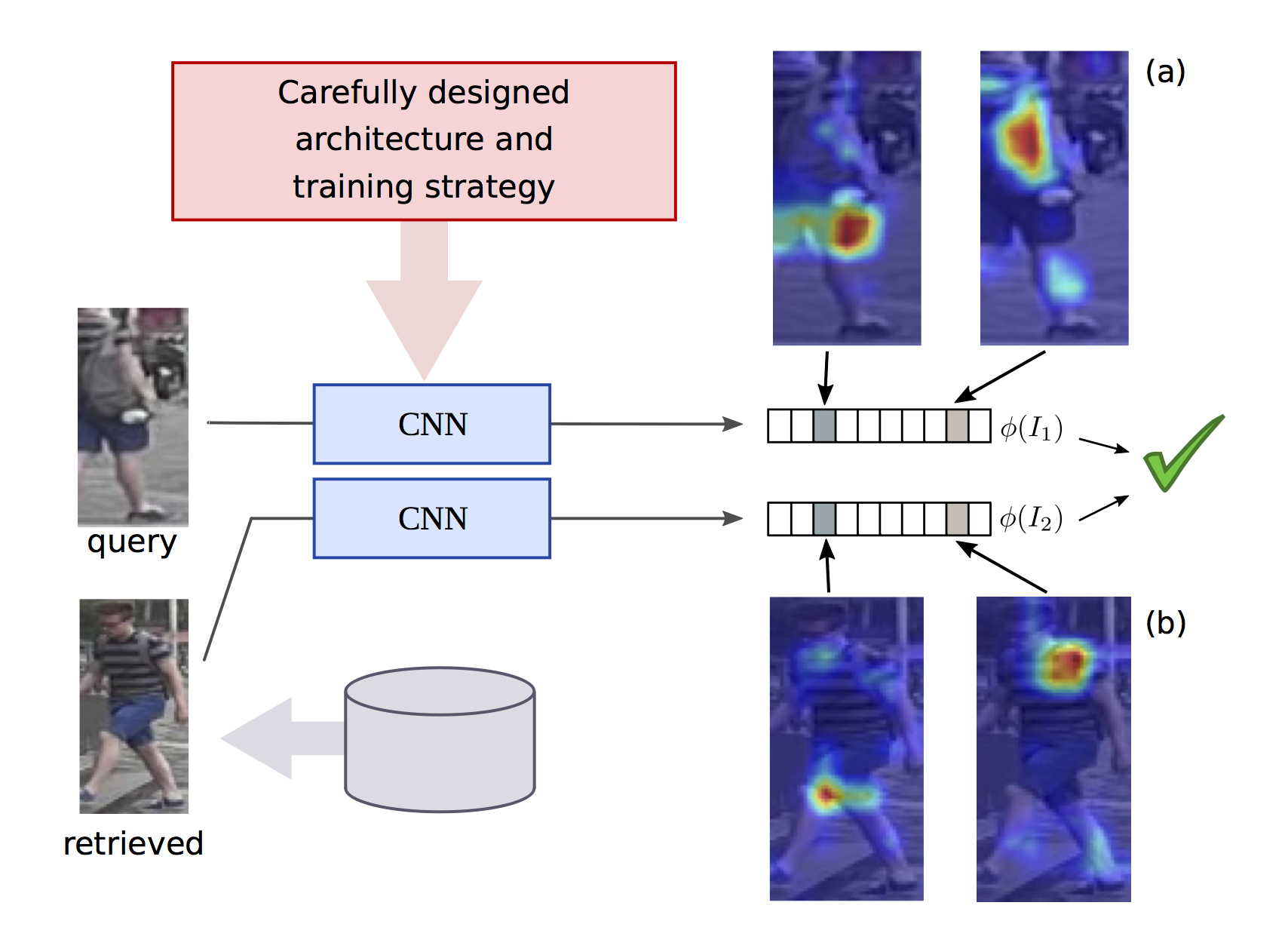}
  \caption{By careful design of our deep architecture and training strategy (Section~\ref{sec:method}), our approach builds global representations that capture the subtle details required for person re-identification by training the embedding dimensions to respond strongly to discriminative regions/concepts such as the backpack or the hem of the shorts. Heatmaps indicate image regions that strongly activate different dimensions of the embedding.
}
  \label{fig:front}
\end{figure}

One potential limitation of using global representations designed for image classification is the absence of any explicit mechanism to tackle the misalignment inherent to human pose variations and person detection errors.
Consequently, many recent works in the literature have explored strategies to alleviate this problem by explicitly aligning body parts between images \cite{su17pose,zhao17spindle}, for example by using pre-trained part or human joint detectors, or by enriching the training set with auxiliary data such as attributes \cite{su16deep}. 

In this work, we adopt a different approach that combines a simple deep network with an appropriate training strategy, and whose design choices were both carefully validated on several datasets.
The result is a simple yet powerful architecture that produces global image representations that, when compared using a dot-product, outperform state-of-the-art person re-identification methods by large margins, including more sophisticated methods that rely on attention models, extra annotations, or explicit alignment.

Our contribution is threefold.
First, we identify a set of key practices to adopt, both for representing images efficiently and for training such representations, when developing person re-ID models (Section \ref{sec:method}). Many of these principles have been adopted in isolation in various related works. However, we show that when applied jointly, significant performance improvements result.
We carefully evaluate different modeling and learning choices that impact performance.
A key conclusion is that curriculum learning is critical for successfully training the image representation and several of our principles reflect this.

Second, our method significantly improves  over previous published results on four standard benchmark datasets for person re-identification (Section~\ref{sec:sota_exp}). For instance, we show an absolute improvement of 8.1\% mAP in the Market-1501 dataset compared with the current state of the art.

Third, we provide a qualitative analysis of the information captured by the visual embedding produced by our architecture. Our analysis illustrates, in particular, the effectiveness of the model in localizing image regions that are critical for re-ID without the need for explicit attention or alignment mechanisms (Section~\ref{sec:qual}). We also show how individual dimensions of the embedding selectively respond to localized semantic regions producing a high similarity between pairs of images from the same person. 

We believe that our approach, which is easy to reproduce\footnote{To aid reproducibility we will release trained models and the evaluation code upon acceptance.}, can serve as a baseline of choice for future improvements in this field of research.

%% file: relatedwork.tex
\begin{figure*}[t!]
  \includegraphics[width=\linewidth]{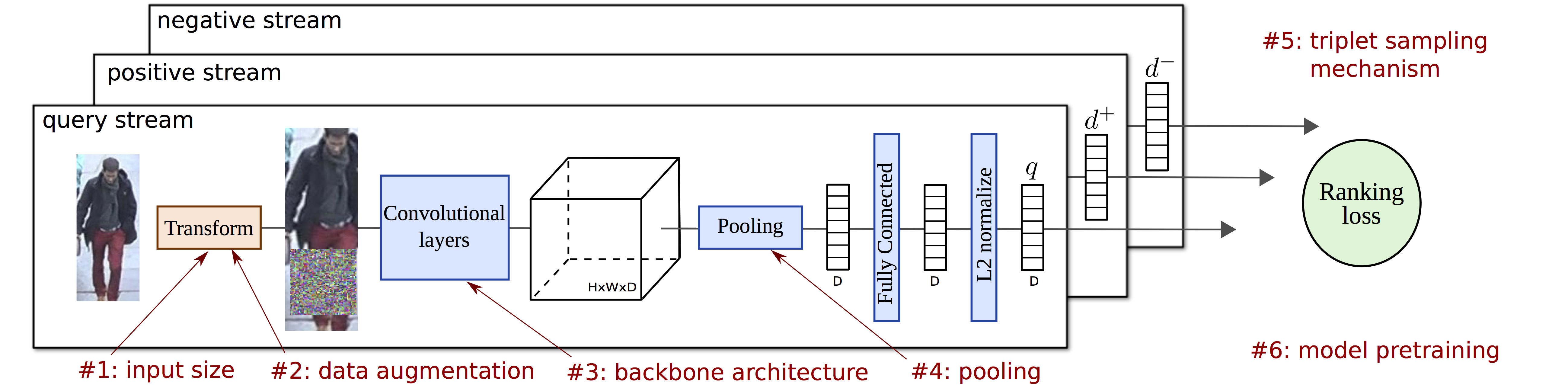}
  \caption{\textbf{Summary of our training approach}. Image triplets are sampled and fed to a three stream Siamese architecture, trained with a ranking loss. Weights of the model are shared across streams. Each stream encompasses an image transformation, convolutional layers, a pooling step, a fully connected layer, and an $\ell_2$-normalization, all these steps being differentiable. In red we show the steps that require a careful design and that we extensively discuss and evaluate in our paper. \label{fig:method}}
\end{figure*}

\section{Related Work}
\label{sec:rw}

A vast literature addresses the person re-identification problem (the reader may refer to \cite{karanam2016comprehensive} for a recent survey).
Traditionally, works on person re-ID sought to improve similarity scores between images \cite{kostinger12large,pedagadi13local,paisitkriangkrai15learning}, usually through metric learning.
These methods typically used color-based histograms as input feature vectors~\cite{mignon12pcca,kostinger12large,pedagadi13local,xiong14person,paisitkriangkrai15learning} due to their discriminative power particularly with respect to clothing, and also to their small memory footprint.
Recent research on person re-identification has mostly focused on end-to-end training of deep architectures. This research has taken two main directions: improving generic deep image representations using sophisticated learning objectives appropriate for person re-identification, or designing task-specific image representations.

\myparagraph{Task-specific learning objectives}
This line of research most often involves proposing loss functions suitable for the re-ID task, and in particular for learning effective similarity metrics.
\cite{zhou17point} proposes a metric to learn similarities between an image and a set of images, as opposed to learning similarities between pairs of images as is typical.
\cite{zhou17efficient} proposes a method to locally modify, in an online manner at test time using only negative examples,  a global similarity metric that was trained offline. \cite{sun17svdnet} added an orthogonality constraint on the final fully-connected layer of a deep network in order to improve the discriminability of the learned features.  
\cite{Zhu2017} proposes to train a re-ID model using as a similarity metric a hybrid of the Euclidean, Cosine and Mahalanobis distances.
\cite{zhang16learning} learns an embedding that aims to project images of the same person into the same point in the embedding space.
\cite{bai17scalable} proposes to learn a method to modify image embeddings such that the learned similarity metric between images is smooth in the underlying image manifold.
\cite{lin17improving} proposes to learn an image embedding for re-ID by training a network to predict both person IDs and attribute labels.

Most recent works use cross-entropy or softmax loss functions for training their person re-identification models.
Others treat person re-ID not as a recognition but rather as a ranking problem, and use losses that are more appropriate for ranking.
For example, the contrastive loss \cite{varior16gated} and the triplet loss or variants thereof \cite{ding15deep, su16deep, hermans17indefense, chen17beyond} have been used to train Siamese architectures. \cite{ding15deep} proposes a scheme to limit the size of triplet batches while still obtaining informative samples, while \cite{chen17beyond} proposes a quadruplet loss, which adds to the triplet loss a term that enforces a margin constraint on the distance between image pairs that are unrelated.
\cite{hermans17indefense} shows that, with appropriate training settings, the triplet loss can outperform more complicated objective functions.
In this work, we propose several good practices that can be viewed as encouraging curriculum learning (\emph{c.f.} section~\ref{sec:curriculum}) that, when combined with the standard triplet loss, lead to large improvements over previous methods which have used varieties of the triplet loss.

\myparagraph{Task-specific representations}
Many works in this line have focused on addressing the alignment problem via use of part detectors, pose estimation, or attention models.
Spatial transformer networks have been used to globally align images \cite{zheng17pan} and to localize salient image regions for finding correspondences~\cite{rahman21person}.
In a similar vein, \cite{zhao17deeply,Liu2017,liu17hydra} use multiple parallel sub-branches which learn, in an unsupervised manner, to consistently attend to different human body parts.
\cite{su17pose} uses a pre-trained pose estimation network to provide explicit part localization, while a similar approach \cite{zhao17spindle} integrates a pose estimation network into their deep re-ID model.
\cite{zheng17pose} uses joint localization to create a new image that contains only the body parts.
Rather than localize parts, \cite{Lin2017} represents images with fixed grids and learns cell correspondences across camera views.
Several works have proposed multi-scale architectures with mechanisms for automatic scale selection \cite{qian17multi} or scale fusion \cite{chen17dpfl}.
\cite{li17learning} combines a multi-scale architecture with unsupervised body part localization using spatial transformer networks.
In Section~\ref{sec:exp}, we compare to such works and show that our learned representation can address alignment and scale variations without using additional scale, human parsing, or attention models.

Other relevant areas of research in re-ID are data scarcity, re-ranking, and end-to-end re-ID.
\cite{zheng17unlabeled} uses GANs to synthesize crops of pedestrians which were used to train a deep re-ID network in a semi-supervised manner.
\cite{zhong17re-ranking} applies \emph{k}-reciprocal nearest neighbor reranking to the re-ID problem.
\cite{liu2017neural, xiao2017joint} both tackle end-to-end re-ID by incorporating person detection into their proposed pipelines.

%% file: method.tex
\section{Learning a global representation for re-ID}
\label{sec:method}

We now describe the design of our deep architecture and our strategy for effectively training it for person re-ID.

\subsection{Architecture design}
\label{sec:architecture}

The architecture of our image representation model in most ways resembles that of standard deep image recognition models.
However, it incorporates several important modifications that  proved beneficial for image retrieval tasks \cite{gordo16deep,Radenovic2016}.
The model contains a backbone convolutional network, pre-trained for image classification, which is used to extract local activation features from input images of an arbitrary size and aspect ratio.
These local features are then max-pooled into a single vector, fed to a fully-connected layer and $\ell_2$-normalized, producing a compact vector whose dimension is independent of the image size. Figure~\ref{fig:method} illustrates these different components and identifies the design choices (\#1 to \#4) that we evaluate in the experimental section (Section~\ref{sec:ablative}).

Different backbone convolutional neural networks, such as ResNet \cite{hermans17indefense}, ResNeXt \cite{xie16resnext}, Inception \cite{szegedy16inception} and Densenet \cite{huang2017densely} can be used interchangeably in our architecture.
In Section~\ref{sec:ablative}, we present results using several flavors of ResNet \cite{hermans17indefense}, and show the influence of the number of convolutional layers on the accuracy of our trained model.

\subsection{Architecture training}

A key aspect of the previously described representation is that all the operations are differentiable. Therefore, all the network weights (\ie from both convolutional and fully-connected layers) can be learned in an end-to-end manner.

\myparagraph{Three-stream Siamese architecture}
To train our representation end-to-end we use a three-stream Siamese architecture in which the weights are shared between all streams. This learning approach has been successfully used for person re-identification \cite{ding15deep,su16deep,hermans17indefense} as well as for different retrieval tasks \cite{gordo16deep,Radenovic2016}.
Since the weights of the convolutional layers and the fully-connected layer are independent of the size of the input image, this Siamese architecture can process images of any size and aspect ratio. 
The three-stream architecture takes image triplets as input, where each triplet contains a query image $I_q$, a positive image $I^+$ (\ie an image of the same person as in the query image), and a negative image $I^-$ (\ie an image of a different person). Each stream produces a compact representation for each image in the triplet, leading to the descriptors $q$, $d^+$ and $d^-$ respectively. 
We then define the ranking triplet loss as
\begin{equation}
  L(I_q,I^+,I^-) = \max (0, m + q^Td^- - q^Td^+),
\end{equation}
where $m$ is a scalar that controls the margin. This loss ensures that the embedding of the positive image $I^+$ is closer to the query image embedding $I_q$ than that of the negative image $I^-$, by at least a margin $m$. 

We now discuss key practices for improved training of our model.

\myparagraph{Image size} Typically, training images are processed in batches and therefore resized to a fixed input size, which leads to distortions. We argue that images should be upscaled to increase the input image size, and that they should not be distorted. To this end, we process triplets sequentially, allowing a different input size for each image and allowing the use of high resolutions images even in the most memory hungry architectures (e.g. ResNet-152 or Densenet). To account for the reduced batch size, we accumulate the gradients of the loss with respect to the parameters of the network for every triplet, and only update the network once we achieve the desired effective batch size. 

\myparagraph{Pretraining} We found it crucial to use pre-trained models with our architecture. First, we follow standard practice and use networks pre-trained on ImageNet \cite{deng09imagenet}. To achieve the highest performance, it was also quite important to perform an additional pre-training step by fine-tuning the model on the training set using a classification loss, \ie to train the model for person identification. We discuss this further in Section~\ref{sec:curriculum} and in the ablative study in Section~\ref{sec:ablative}.

\begin{figure}[t!]
  \centering
%\fbox
\fcolorbox{black}{silver}{
  \begin{minipage}[c]{0.95\linewidth}
    \footnotesize
    Good practices for person re-ID
      \begin{itemize}
      \item Pre-training for identity classification
      \item Sufficiently large image resolution
      \item State-of-the-art base architecture % (\eg ResNet 100)
      \item Hard triplet mining
      \item Dataset augmentation with difficult examples%: we use image cut-out to achieve this.
      \end{itemize}
  \end{minipage}
}
\caption{Summary of good practices for building a powerful representation for person re-identification.
\label{fig:recipe}}
\end{figure}

\myparagraph{Data augmentation}
 
To augment the dataset we adopt an image ``cut-out'' strategy, which consists of adding random noise to random-sized regions of the image.
We progressively increase the maximum size of these regions during training, progressively producing more difficult examples.
This strategy improves the results because it serves two purposes: it is a data augmentation scheme that directly targets robustness to occlusion and it allows for model regularization by acting as a ``drop-out'' mechanism at the image level. As a result, this strategy avoids the over-fitting inherent to the small size of the training set and significantly improves the results.
We also considered standard augmentation strategies such as image flipping and cropping \cite{simonyan16vgg} but found no added improvement, as we show in Section~\ref{sec:ablative}.

\myparagraph{Hard Triplet Mining} Finally, mining \emph{hard} triplets is crucial for learning. 
As already argued in \cite{hermans17indefense, harwood17mining, wu17sampling}, when applied naively, training with a triplet loss can lead to underwhelming results.
Here we follow the hard triplet mining strategy introduced in \cite{gordo2017end}.
First, we extract the features for a set of $N$ randomly selected examples using the current model and compute the loss of all possible triplets. Then, to select triplets, we randomly select an image as a query and randomly pick a triplet for that query from among the 25 triplets with the largest loss. To accelerate the process, we only extract a new set of random examples after the model has been updated $k$ times with the desired batch size $b$. This is a simple and effective strategy which yields good model convergence and final accuracy, although other hard triplet mining strategies \cite{harwood17mining, wu17sampling} could also be considered.

\subsection{Curriculum learning for re-ID}
\label{sec:curriculum}
Similarly to humans, who learn a set of concepts more easily when the concepts to be learned are presented by increasing degree of complexity, it has been shown that curriculum learning has a positive impact on the speed and quality of the convergence of deep neural networks \cite{bengio09curriculum}.
We adopt this learning strategy in our approach.
In particular, three of our design principles described in this section aim to progressively increase the difficulty of the task being learned by our model. 
First, our hard-negative mining strategy samples triplets that increase in difficulty as learning continues.
Second, our pre-training strategy first trains our model to solve the task of person ID classification (which requires the model to first recognize individuals within a closed set of possible IDs) before training it for the more challenging task of re-identifying persons. Third we observed that when training with cut-out, we achieve best results when the percentage of the image that is replaced by noise progressively increases.
We believe that this general training principle is crucial to our results (reported in Section~\ref{sec:sota_exp}).

Figure~\ref{fig:recipe} summarizes the good practices that we propose for both designing and training a deep architecture for person re-identification.

%% file: experiments.tex
\section{Experiments}
\label{sec:exp}

\newcolumntype{K}[1]{>{\centering\arraybackslash}p{#1}}

%%%%%%%%%%%%%%%%%%%%%%%%%%%%%%%%%%%%%%%%%%%%%%%%%%%%%%%%%%%%%%%%%%%%%%%%%%%%%%%%%%%%%%%%%%%%%%%%%%
\subsection{Experimental details}
\label{sec:details}
%%%%%%%%%%%%%%%%%%%%%%%%%%%%%%%%%%%%%%%%%%%%%%%%%%%%%%%%%%%%%%%%%%%%%%%%%%%%%%%%%%%%%%%%%%%%%%%%%%

\myparagraph{Datasets} We consider four datasets for evaluation.

\noindent The {\bf Market-1501} dataset~\cite{zheng15scalable} (Market) is a standard person re-ID benchmark with images from 6 cameras of different resolutions. DPM detections \cite{felzenszwalb2010object} were annotated as containing one of the 1,501 identities, among which 751 are used for training and 750 for testing. The training set contains 12,936 images with 3,368 query images. The gallery set is composed of images from the 750 test identities and of distractor images, 19,732 images in total.
There are two possible evaluation scenarios for this database, one using a single query image and one with multiple query images.

\noindent The {\bf MARS} dataset~\cite{zheng16mars} is an extension of Market that targets the retrieval of gallery tracklets (\ie sequences of images) rather than individual images. It contains 1,261 identities, divided into a training (631 IDs) and a test (630 IDs) set. The total number of images is 1,067,516, among which 518,000 are used for training and the remainder for testing.

\noindent The {\bf DukeMTMC-reID} dataset~\cite{zheng17unlabeled} (Duke) was created by manually annotating pedestrian bounding boxes every 120 frames of the videos from 8 cameras of the original DukeMTMC dataset~\cite{ristani16performance}. It contains 16,522 images of 702 identities in the training set, and 702 identities, 2,228 query and 17,661 gallery images in the test set.

\noindent The {\bf Person Search} dataset \cite{xiao2017joint} (PS) differs from the previous three as it was created from images collected by hand-held cameras and frames from movies and TV dramas. It can therefore be used to evaluate person re-identification in a setting that doesn't involve a known camera network. It contains 18,184 images of 8,432 identities, among which 5,532 identities and 11,206 images are used for training, and 2,900 identities and 6,978 images are used for testing.

%%%%%%%%%%%%%%%%%%%%%%%%%%%%%%%%%%%%%%%%%%%%%%%%%%%%%%%%%%%%%%%%%%%%%%%%%%%%%%%%%%%%%%%%%%%%%%%%%%
  \begin{table}[t!]
    \small
    \centering
    \begin{tabular}{cccK{1.5cm}K{1.5cm}}
      \toprule
      flip & crop  & cut-out & Market & Duke \\
      \midrule
      -     & -  &     -       & 75.9 & 69.6  \\
      \cmark  & - &      -       & 77.2 & 69.7 \\
      -  &  \cmark & -            & 76.8 & 69.4 \\
      -  & -  & \cmark       & \textbf{81.2} & \textbf{72.9} \\
      
      \cmark  & \cmark  & \cmark & \textbf{81.2} & \textbf{72.9} \\
      \bottomrule
    \end{tabular}
    \caption{{\bf Impact of different data augmentation strategies}. We report mean average precision (mAP) on Market and Duke.\label{tab:dataaugmentation}}
\end{table}
%%%%%%%%%%%%%%%%%%%%%%%%%%%%%%%%%%%%%%%%%%%%%%%%%%%%%%%%%%%%%%%%%%%%%%%%%%%%%%%%%%%%%%%%%%%%%%%%%%

%%%%%%%%%%%%%%%%%%%%%%%%%%%%%%%%%%%%%%%%%%%%%%%%%%%%%%%%%%%%%%%%%%%%%%%%%%%%%%%%%%%%%%%%%%%%%%%%%%
  \begin{table}[t!]
  \small
  \centering
  \begin{tabular}{cK{1.5cm}K{1.5cm}}
    \toprule
    Largest dimension & Market & Duke \\
    \midrule 
    256 pixels  & 78.2 & 69.2  \\
    416 pixels  & 81.2 & 72.9  \\
    640 pixels  & 81.2 & 73.1  \\
  \bottomrule
  \end{tabular}
 \caption{\textbf{Impact of the input image size}. We report mean average precision (mAP) on Market and Duke. \label{tab:inputsize}}
  \end{table}
%%%%%%%%%%%%%%%%%%%%%%%%%%%%%%%%%%%%%%%%%%%%%%%%%%%%%%%%%%%%%%%%%%%%%%%%%%%%%%%%%%%%%%%%%%%%%%%%%%

\myparagraph{Evaluation} We follow standard procedure for all datasets and report the mean average precision over all queries (mAP) and the cumulative matching curve (CMC) at rank-1 and rank-5 using the evaluation codes provided.

\myparagraph{Training details}
As mentioned in Section~\ref{sec:architecture}, for the convolutional part of our network we evaluate different flavors of ResNet \cite{he16deep}, concretely ResNet-50, ResNet-101 and ResNet-152 (we study their impact in the following section). For all of them, we start with the publicly available pre-trained model on ImageNet, and fine-tune the weights of the convolutional layers for person identification in the training set of the specific dataset. To do this, we follow standard practice and extract random-sized crops and then resize them to $224 \times 224$ pixels. We train with stochastic gradient descent (SGD) with momentum of $0.9$, weight decay of $5 \cdot 10^{-5}$, a batch size of $128$, and an initial learning rate of $10^{-2}$, which we decrease to $10^{-4}$. We use the weights of this pre-trained network for the convolutional layers of our architecture and we randomly initialize the fully-connected layer, whose output we set to 2,048 dimensions. We then train the ranking network using our Siamese architecture with input images of variable size, while fixing the largest side to $M$ pixels (whose influence we also study in the following section). We use again SGD with a batch size of $64$ and an initial learning rate of $10^{-3}$, which we decrease using a logarithmic rate that halves the learning rate every $512$ iterations. We observe in all our experiments that the model converges after approximately 4,096 iterations.
For the hard triplet mining we set the number of random examples to $N=5,000$ and the number of updates to $k=16$.
We set the margin of the triplet loss to $m=0.1$.
Exactly the same training settings were used across all four datasets.

 \input{magic_table}

%%%%%%%%%%%%%%%%%%%%%%%%%%%%%%%%%%%%%%%%%%%%%%%%%%%%%%%%%%%%%%%%%%%%%%%%%%%%%%%%%%%%%%%%%%%%%%%%%%
\subsection{Ablative study}
%%%%%%%%%%%%%%%%%%%%%%%%%%%%%%%%%%%%%%%%%%%%%%%%%%%%%%%%%%%%%%%%%%%%%%%%%%%%%%%%%%%%%%%%%%%%%%%%%%
\label{sec:ablative}

In this section we evaluate key design choices in our architecture and training strategy that relate to the good practices we propose in Figure~\ref{fig:recipe}.  

\myparagraph{Image transformation}
We first focus on data augmentation (\#2 in Figure~\ref{fig:method}). As discussed in Section~\ref{sec:method}, we apply different transformations to the images at training time, namely flips, crops and cut-outs. Here we study how each transformation impacts the final results, reported in Table~\ref{tab:dataaugmentation}.
We observe that cut-out has a very strong impact on the performance and renders the other two data augmentation schemes superfluous. We believe that this is because cut-out makes our representation much more robust to occlusion, and also avoids over-fitting on such little training data. 

Second, we consider the impact of the size of the input image (\#1). Images from the Market dataset have a fixed size of $256 \times 128$, while images from Duke have a variable size, with $256 \times 128$ pixels on average. 
In our experiments, we rescale images so that the largest image dimension is either 256, 416, or 640 pixels, without distorting the aspect ratio. We report results in Table~\ref{tab:inputsize} and observe that using a sufficiently large resolution is key to achieving the best performance. Increasing the resolution from 256 to 416 improves mAP by 3\%, while increasing it further to 640 pixels shows negligible improvement. We set the input size to 416 pixels for the rest of this paper.

\input{table_sota}

\myparagraph{Pooling}
Table~\ref{tab:magic} (a) compares two pooling strategies (\#4) over the feature map produced by the convolutional layers. As we see that max pooling performs better than average pooling on both datasets, we use it for the rest of this paper. 
 
\myparagraph{Backbone architecture}
 Table~\ref{tab:magic} (b) compares different architectures for the convolutional backbone of our network (\#3). Results show that using ResNet-101 significantly improves the results compared with using ResNet-50 (about +5 mAP for both datasets). The more memory hungry ResNet-152 only marginally improves the results.

\myparagraph{Fine-tuning for classification}
Table~\ref{tab:magic} (c) shows the importance of fine-tuning the convolutional layers for the identity classification task before using the ranking loss to adjust the weights of the whole network (\#6). As discussed in Section~\ref{sec:curriculum}, training the model on tasks of increasing difficulty is highly beneficial.

%%%%%%%%%%%%%%%%%%%%%%%%%%%%%%%%%%%%%%%%%%%%%%%%%%%%%%%%%%%%%%%%%%%%%%%%%%%%%%%%%%%%%%%%%%%%%%%%%%
\subsection{Comparison with the state of the art}\label{sec:sota_exp}
%%%%%%%%%%%%%%%%%%%%%%%%%%%%%%%%%%%%%%%%%%%%%%%%%%%%%%%%%%%%%%%%%%%%%%%%%%%%%%%%%%%%%%%%%%%%%%%%%%

Table~\ref{tab:marsket_results} compares our approach to the state of the art. Our method consistently outperforms all methods by large margins on all 4 re-ID datasets and all metrics.
In particular, we achieve a mAP of 81.2\% on Market, an 8.1\% absolute improvement compared with the best published results \cite{chen17dpfl}. %and a 12.1\% improvement over the next-best method of \cite{hermans17indefense}, which is unpublished work. 
We also outperform \cite{hermans17indefense} by 12.0\% mAP on MARS.
On the Duke dataset, we achieve a mAP of 72.8\%, outperforming the previous best reported mAP \cite{chen17dpfl} by 12.2\%.
It is also important to note that our approach using ResNet-50, reported in Table~\ref{tab:magic} b), still outperforms prior art by a significant margin, showing that all of our design choices play a crucial role, not only the backbone architecture.
We also report the performance of our method with standard re-ranking\footnote{We expand both the query and the dataset by averaging the representation of the first 5 and 10 closest neighbors, respectively.} and we again see large improvements with respect to prior art that uses re-ranking, across all datasets and metrics. For example, for Market, we achieve a mAP of 90\%, 8.9\% above the best previously-reported mAP from \cite{hermans17indefense}. 

Looking closely at the approaches that report results on these datasets, we first note that our approach outperforms all recent methods that also use a variant of the triplet loss and hard triplet mining \cite{hermans17indefense, zhao17deeply}. As we show in this section, combining these key principles with the others mentioned in Figure~\ref{fig:recipe} is crucial for effective training of our image representation for Re-ID. It is also worth emphasizing that our approach even outperforms recent works that propose complex models for aligning images based on attributes \cite{lin17improving} or body parts via pose estimation \cite{zheng17pose}, part detection \cite{li17learning, zhao17spindle} or attention modules \cite{zhao17deeply}, most of which require extra resources such as annotations or pre-trained detectors. As we discuss in the next section, our model is able to discriminate body regions without such additional architectural modules.

We also report results for the Person Search dataset in last column of Table~\ref{tab:marsket_results}. This dataset differs from traditional re-ID datasets in that the different views of each person do not correspond to different cameras in a network.
Nevertheless, our approach performs quite well in this different scenario, achieving a mAP of 92.6\%, which is a 14.7\% absolute improvement over the previous best reported result \cite{xiao2017joint}. This shows the generality of our approach. 

%%%%%%%%%%%%%%%%%%%%%%%%%%%%%%%%%%%%%%%%%%%%%%%%%%%%%%%%%%%%%%%%%%%%%%%%%%%%%%%%%%%%%%%%%%%%%%%%%%
\subsection{Qualitative analysis}\label{sec:qual}
%%%%%%%%%%%%%%%%%%%%%%%%%%%%%%%%%%%%%%%%%%%%%%%%%%%%%%%%%%%%%%%%%%%%%%%%%%%%%%%%%%%%%%%%%%%%%%%%%%

In this section we perform a detailed analysis of our trained model's performance and induction biases.

\begin{figure}[t!]
  \includegraphics[width=\linewidth]{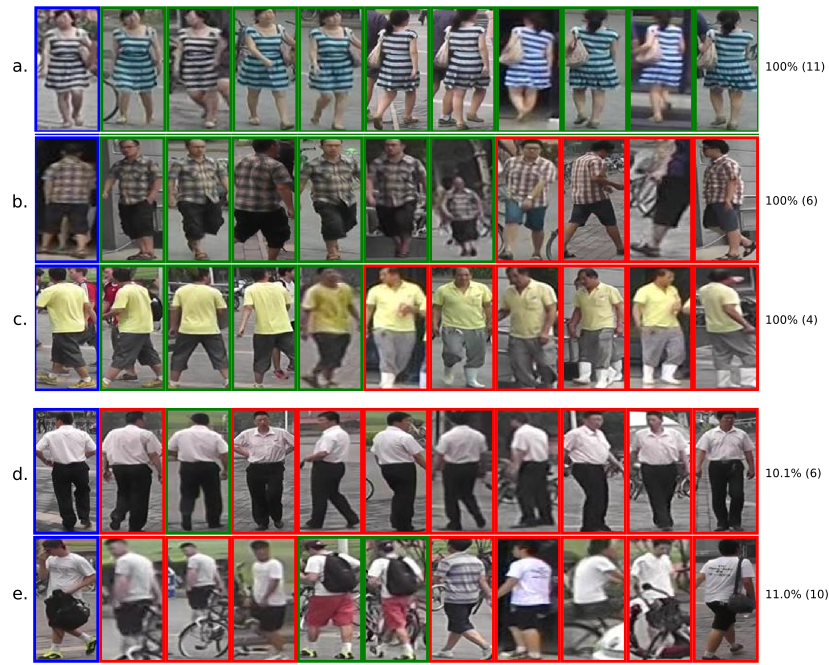}
  \caption{For several queries from Market, we show the first 10 retrieved images together with the mAP and the number of relevant images (in brackets) of that query. Green (resp. red) outlines images that are relevant (resp. non-relevant) to the query.}
  \label{fig:good_bad}
\end{figure}

\myparagraph{Re-identification examples}
In Figure~\ref{fig:good_bad}, we show good results (top) and failure cases (bottom) for several query images from the Market dataset. We see that our method is able to correctly re-identify persons despite pose changes or strong scale variations. We observe that failure cases are mostly due to confusions between two people that are extremely difficult to differentiate even for a human annotator, or to unusual settings (for instance the person holding a backpack in front of him as in e.).

\begin{figure*}[t!]
  \includegraphics[width=\linewidth]{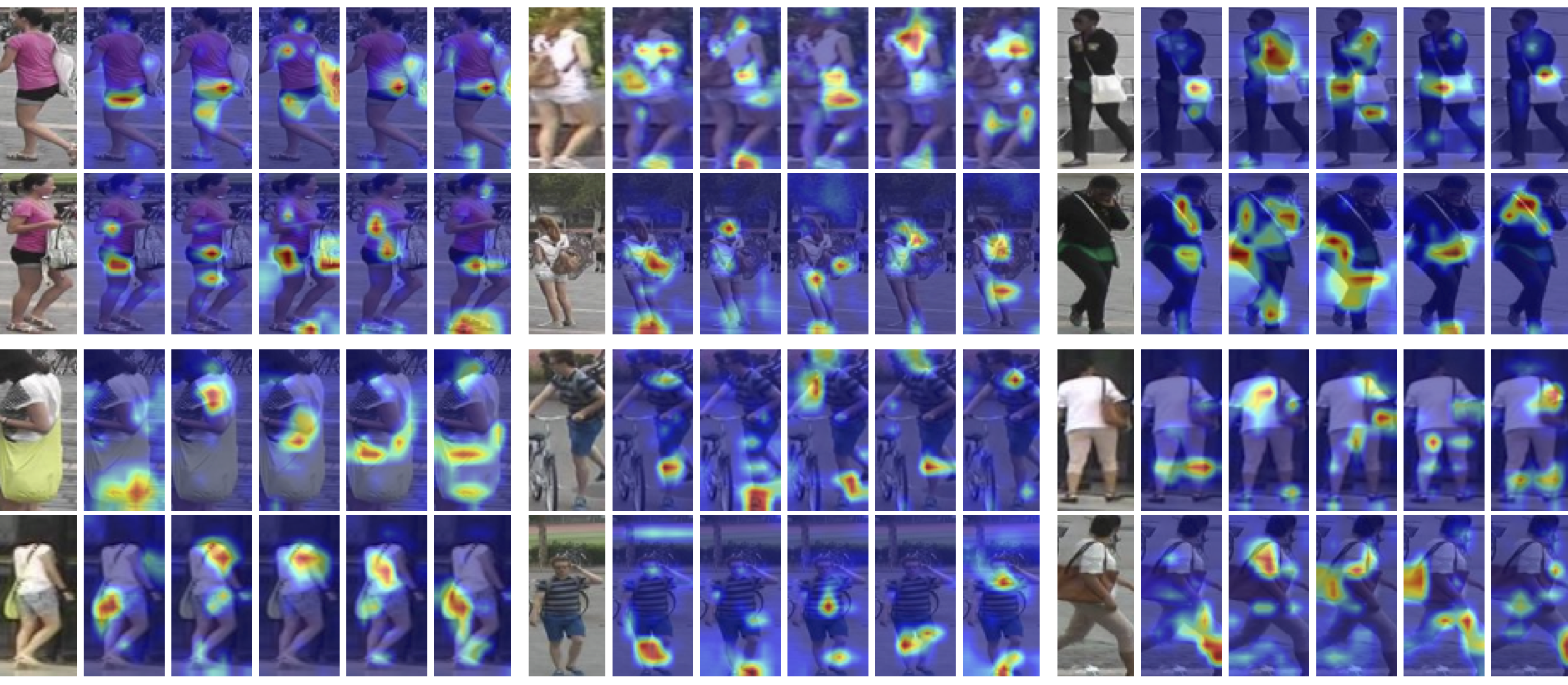}
  \caption{{\bf Matching regions} For pairs of matching images, we show maps for the top 5 dimensions that contribute most to the similarity.}
  \label{fig:pairwise}
\end{figure*}

\myparagraph{Localized responses and clothing landmark detection}
In Section~\ref{sec:method}, we argued that, using our proposed approach, we obtain an embedding that captures invariance properties useful for re-ID. To qualitatively analyze this invariance, we use Grad-Cam~\cite{selvaraju17gradcam}, a method for highlighting the discriminative regions that CNN-based models activate to predict visual concepts. This is done by using the gradients of these concepts flowing into the final convolutional layer. Similar to~\cite{gordo2017beyond}, given two images, we select the 5 dimensions that contribute the most to the dot-product similarity between their representations. Then, for each image, we propagate the gradients of these 5 dimensions individually, and visualize their activations in the last convolutional layer of our architecture. In Figure~\ref{fig:pairwise}, we show several image pairs and their respective activations for the top 5 dimensions. 

We first note that each of these output dimensions are activated by fairly \textit{localized image regions} and that the dimensions often reinforce one-another in that image pairs are often activated by the same region. This suggests that the similarity score is strongly influenced by localized image content. 
Interestingly, these localized regions tend to contain body regions that can inform on the type of clothing being worn. Examples in the figure include focus on the hem of a pair of shorts, the collar of a shirt, and the edge of a sleeve. Therefore, rather than focusing on aligning human body joints, the model appears to make decisions based on \textit{attributes of clothing} such as the length of a pair of pants or of a shirt's sleeves. This type of information has been leveraged explicitly for retrieval using the idea of ``fashion landmarks'', as described in \cite{liu2016deepfashion}.
Finally, we observe that some of the paired responses go \textit{beyond appearance similarity} and respond to each other at a more abstract and semantic level. For instance, in the top right pair the strong response of the first dimension to the bag in the first image seems to pair with the response to the strap of the bag in the second image, the bag itself being occluded (see also the backpack response of Figure~\ref{fig:front} as an other example).

\myparagraph{Implicit attention}
We now qualitatively examine which parts of the images are highly influential, independently of the images they are matched with.
To do so, given an image and its embedding, we select the first 50 dimensions with the strongest activations. We then propagate and accumulate the gradients of these dimensions, again using Grad-Cam \cite{selvaraju17gradcam}, and visualize their activations in the last convolutional layer in our architecture.
As a result, we obtain a visualization that highlights parts of the images that, \emph{a priori}, will have the most impact on the final results. This can be seen as a visualization of the implicit attention mechanism that is at play in our learned embedding.

\begin{figure}[t!]
\centering
  \includegraphics[width=\linewidth]{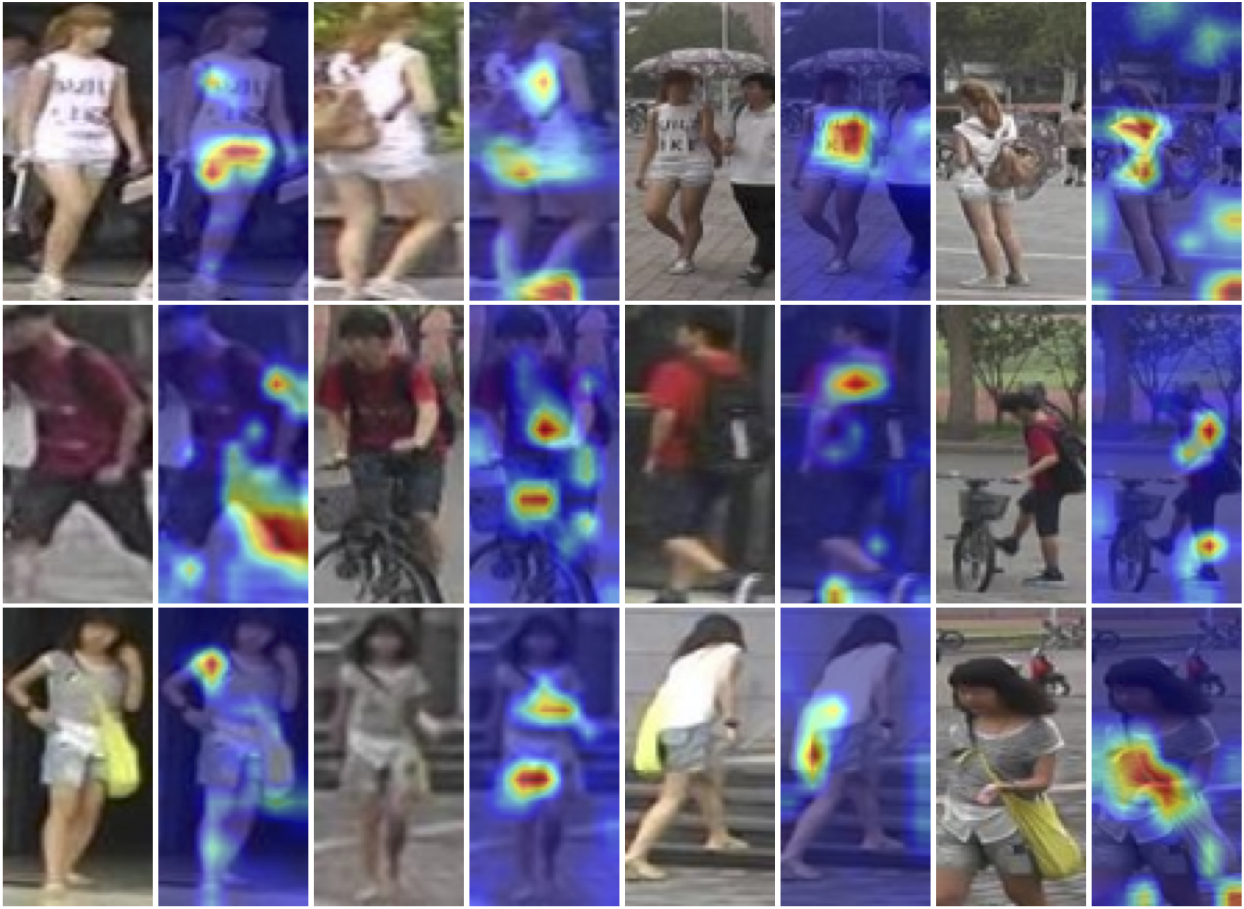}
  \caption{{\bf Implicit attention} 
  We highlight regions that correspond to the most highly-activated dimensions of the final descriptor. 
  They focus on unique attributes, such as backpacks, bags, or shoes.\label{fig:attention}}
\end{figure}

We show such \textit{implicit attention masks} in Figure~\ref{fig:attention} across several images of the same person, for three different persons. We first observe that the model attends to regions known to drive attention in human vision, such as high-resolution text \cite{cerf2009faces} (e.g. in rows 1 and 2). We also note that our model shows properties of contextual attention, particularly when image regions become occluded. For example, when the man in the second row faces the camera, text on his t-shirt and the hem of his pants are attended to. However, when his back or side is to the camera, the model focuses more intently on the straps of his backpack.

%%%%%%%%%%%%%%%%%%%%%%%%%%%%%%%%%%%%%%%%%%%%%%%%%%%%%%%%%%%%%%%%%%%%%%%%%%%%%%%%%%%%%%%%%%%%%%%%%%
\subsection{Re-ID in the presence of noise}
%%%%%%%%%%%%%%%%%%%%%%%%%%%%%%%%%%%%%%%%%%%%%%%%%%%%%%%%%%%%%%%%%%%%%%%%%%%%%%%%%%%%%%%%%%%%%%%%%%

To test the robustness of our model, we evaluate it in the presence of noise using Market+500K \cite{zheng15scalable}, an extension of the Market dataset that contains an additional set of 500K distractors.
To generate these distractors, the authors first collected ground-truth bounding boxes for persons in the images. They then computed the IoU between each predicted bounding box and ground-truth bounding box for a given image.
A detection was labeled a distractor if its IoU with all ground-truth annotations was lower than 20\%.

\begin{figure}[t!]
    \centering
  \includegraphics[width=0.85\linewidth]{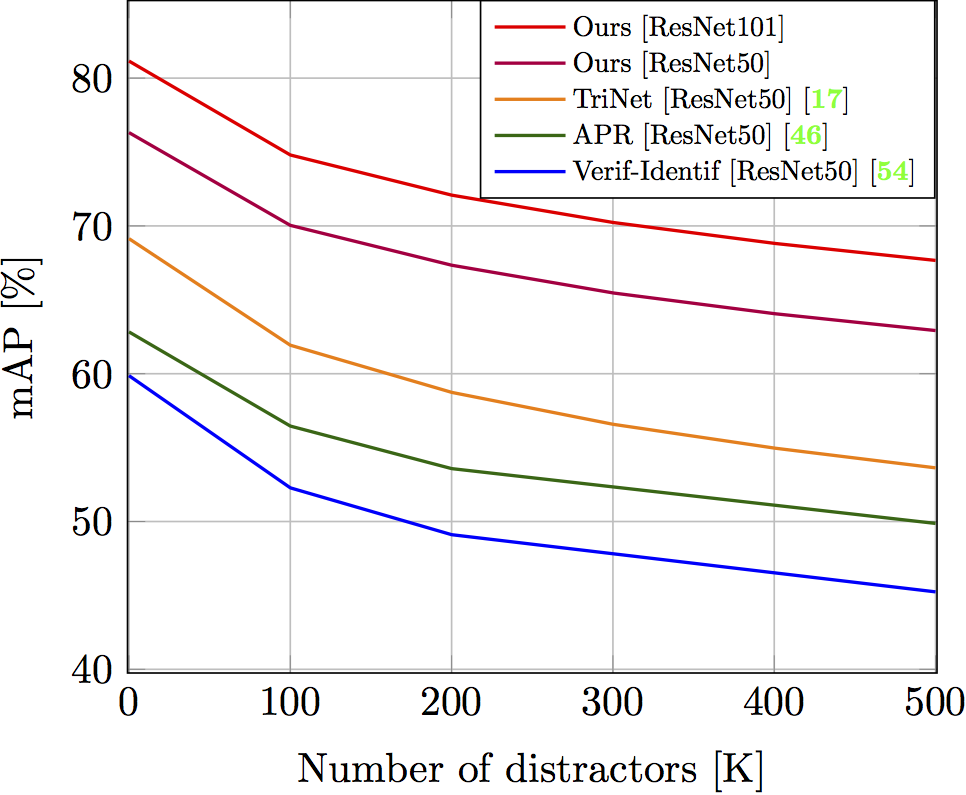}
  \caption{Performance comparison in the presence of distractors.\label{fig:noise}}
  \vspace{-4mm}
\end{figure}

We evaluate our ResNet-50- and ResNet-100-based models, trained on Market, on this expanded dataset, while increasing the number of distractors from 0 to 500K.
We selected distractors by randomly choosing them from the distractor set and adding them to the gallery set.
Both models significantly outperform the current state-of-the-art results in the presence of this noise, as presented in Figure \ref{fig:noise}. Note that our best model, with 500K added distractors, performs on par with \cite{hermans17indefense}'s performance with 0 added distractors.

%% file: magic_table.tex
\begin{table}[t!]
  \small
  \centering
  \begin{tabular}{llcc}
    \toprule
    & & Market & Duke \\
    \midrule
    \multirow{2}{*}{a) pooling strategy} &
      average & 80.1 & 71.4 \\
    & max     & \textbf{81.2} & \textbf{72.9} \\
    \midrule
    \multirow{3}{*}{b) backbone architecture} &
      ResNet-50  & 76.3 & 67.6  \\
    & ResNet-101 & 81.2 & 72.9  \\
    & ResNet-152 & \textbf{81.4} & \textbf{74.0}    \\
    \midrule
    \multirow{2}{*}{c) pretraining for class.} &
      no  & 77.1 & 71.1    \\
    & yes & \textbf{81.2} & \textbf{72.9}  \\
   \bottomrule
  \end{tabular}
  \caption{Top (a): influence of the \textbf{pooling strategy}. Middle (b): results for different \textbf{backbone architectures}. Bottom (c): influence of \textbf{pretraining the network for classification} before considering the triplet loss. We report mAP for Market and Duke. \label{tab:magic}}
\end{table}

%% file: table_sota.tex
\begin{table*}[t]
  \footnotesize
  
\setlength{\tabcolsep}{1pt}
\setlength{\extrarowheight}{5pt}
\renewcommand{\arraystretch}{0.75}
\centering

\begin{tabularx}{\textwidth}{p{2.5cm}cCp{1.3pt}Cp{1pt}Cp{4pt}Cp{1pt}Cp{1pt}Cp{4pt}Cp{1pt}Cp{1pt}Cp{4pt}Cp{1pt}Cp{1pt}Cp{1pt}C}
\toprule
& {\hspace*{5pt}\rotatebox{90}{\hspace*{-10pt}Type}} & \multicolumn{5}{c}{Market-1501 SQ} && \multicolumn{5}{c}{Market-1501 MQ} && \multicolumn{5}{c}{MARS} && \multicolumn{5}{c}{Duke-reID} && PS\\
\cmidrule[0.5pt]{3-7} \cmidrule[0.5pt]{9-13} \cmidrule[0.5pt]{15-19} \cmidrule[0.5pt]{21-25} \cmidrule[0.5pt]{27-27}
 && mAP && rank-1 && rank-5 && mAP && rank-1 && rank-5 && mAP && rank-1 && rank-5 && mAP && rank-1 && rank-5 && mAP\\
\midrule[0.5pt]
 MG~\cite{varior16gated}		& - & 39.6 && 65.9 && - && 48.4 && 76.0 && - && - && - && - && - && - && - && -  \\ % Co
 CRAFT~\cite{chen17person}		& - & 45.5 && 71.8 && - && 54.3 && 79.7 && - && - && - && - && - && - && - && -   \\ %? 
 SpindleNet~\cite{zhao17spindle}		& P & - && 76.9	&& 91.5 && - && - && - && - && - && - && - && - && - && -  \\ %S
 Zheng \etal ~\cite{zheng16mars}       		& - & - && - && - && - && - && - && 49.3 && 68.3 && 82.6 && - && - && - && -  \\ % F
 Part-Aligned~\cite{zhao17deeply}     	& A & 63.4 && 81.0 && 92.0 && - && - && - && - && - && - && - && - && -&& -   \\ %T
 OL-MANS~\cite{zhou17efficient} 	& - & - && 60.7 && - && - && 66.8 && - && - && - && - && - && - && - && -  \\ % 
 HydraPlus-Net~\cite{liu17hydra} 	& A & - && 76.9 && 91.3 && - && - && - && - && - && - && - && - && - && -  \\ % 
  MSCAN~\cite{li17learning} 		& P & 57.5 && 80.3 && - && 66.7 && 86.8 && - && 66.4 && \textbf{83.0} && \textbf{93.7} && - && - && - && -  \\ % S
  OIM~\cite{xiao2017joint}     		& - & - && 82.1 && - && - && - && - && - && - && - && - && 68.1 && - && \textbf{77.9}  \\ % F
 PDC~\cite{su17pose} 	                & P & 63.4 && 84.1 && 92.7 && - && - && - && - && - && - && - && - && - && -  \\ % 
 Verif-Identif.~\cite{zheng2016discriminatively} 	& - & 59.9 && 79.5 && - && 70.3 && 85.8 && - && - && - && - && 49.3 && 68.9 && - && -  \\ % C
  LSRO~\cite{zheng17unlabeled} 	& - & 66.1 && 84.0 && - && 76.1 && 88.4 && - && - && - && - && 47.1 && 67.7 && - && -  \\ % C
  SVDNet~\cite{sun17svdnet} 		& - & 62.1 && 82.3 && 92.3 && - && - && - && - && - && - && 56.8 && 76.7 && \textbf{86.4} && -  \\ %C
 SSM~\cite{bai17scalable}		& - & 68.8 && 82.2 && - && 76.2 && 88.2 && - && - && - && - && - && - && - && -  \\ % C
    DPFL~\cite{chen17dpfl} 		& - & \textbf{73.1} && \textbf{88.9} && 92.3 && - && - && - && - && - && - && \textbf{60.6} && \textbf{79.2} && - && -  \\ %C
DML\cite{zhang17deep}$^*$		& - & 68.8 && 87.7 && - && \textbf{77.1} && \textbf{91.7} && - && - && - && - && - && - && -&& -   \\ %C
APR~\cite{lin17improving}$^*$	& At & 64.7 && 84.3 && 93.2 && - && - && - && - && - && - && 51.9 && 70.7 && - && -  \\ % C
 PAN~\cite{zheng17pan}$^*$ & A & 63.3 && 82.8 && 93.5 && - && - && - && - && - && - && 51.5 && 71.6 && 83.9 && -  \\ % 
 PBF~\cite{zheng17pose}$^*$		& P & 56.0 && 79.3 && 90.8 && - && - && - && - && - && - && - && - && - && -  \\ % S
TriNet~\cite{hermans17indefense}$^*$	& - & 69.1 && 84.9 && \textbf{94.2} && 76.4 && 90.5 && \textbf{96.3} && \textbf{67.7} && 79.8 && 91.4 && - && - && - && -  \\

 \midrule[0.5pt]
 Ours                    & - & \red{81.2} && \red{92.2} && \red{97.9} && \red{87.3} && \red{94.7} && \red{98.6} && \red{79.7} && \red{85.8} && \red{96.5} && \red{72.8} && \red{85.2}  && \red{93.9} && \red{92.6} \\
 Improvement     & - & +8.1 && +3.3 && +3.7 && +10.2 && +3.0 && +2.3 && +12.0 && +2.8 && +2.8 && +12.2 && +6.0 && +7.5 && +14.7  \\
 \midrule[0.5pt]
 \midrule[0.5pt]
Re-ranking~\cite{zhong17re-ranking}	& - & 63.6 && 77.1 && - && - && - && - && 68.4 && 73.9 && - && - && - && - && -  \\ 
TriNet (re-rank)~\cite{hermans17indefense}$^*$& - & 81.1 && 86.7 && 93.4 && 87.2 && 91.8 && 95.8 && 77.4 && 81.2 && 90.8 && - && - && -&& -   \\
\midrule[0.5pt]
Ours (re-rank)      & - & \red{90.0} && \red{93.0} && \red{95.9} && \red{91.2} && \red{94.2} && \red{96.9} && \red{85.7} && \red{87.2} && \red{94.9} && \red{85.6} && \red{89.4} && \red{93.6} && -  \\

\bottomrule[1pt]
\end{tabularx}
\caption{Comparison with state of the art methods on the Market-1501, MARS, Duke-reID and Person Search datasets. The ``Type" column indicates methods that include the following additional components: a part-based representation (P) with extra annotations, an attention mechanism (A), or attribute annotations at train time (At). Bold numbers show the current state  of the art, while red numbers correspond to the best number overall. $^*$ indicates methods published only in \emph{arXiv}.}
\label{tab:marsket_results}
\end{table*}

%% file: conclusions.tex
\section{Conclusions}
\label{sec:ccl}

In this paper, we have proposed a set of good practices for designing and training an efficient and effective image representation model for the task of person re-identification.
We showed through extensive experiments that our model outperforms all state-of-the-art approaches for this task by large margins, across four datasets and three metrics.
We believe that our proposed approach can serve as a useful baseline for future contributions to the field.